\documentclass{Interspeech2024}
\usepackage{algorithm}
\usepackage{pifont}
\usepackage{multirow}
\usepackage{booktabs} 
\usepackage{threeparttable} 
\usepackage[noend]{algpseudocode}

\interspeechcameraready

\title{LASER: Learning by Aligning Self-supervised Representations of Speech for Improving Content-related Tasks}

\name{Amit Meghanani, Thomas Hain}

\email{\{ameghanani1,t.hain\}@sheffield.ac.uk}
\address{Speech and Hearing Research Group \\ Department of Computer Science, The University of Sheffield, United Kingdom}

\begin{document}

\maketitle

\begin{abstract}
Self-supervised learning (SSL)-based speech models are extensively used for full-stack speech processing. However, it has been observed that improving SSL-based speech representations using unlabeled speech for content-related tasks is challenging and computationally expensive. Recent attempts have been made to address this issue with cost-effective self-supervised fine-tuning (SSFT) approaches. Continuing in this direction, a cost-effective SSFT method named ``LASER: \textbf{L}earning by \textbf{A}ligning \textbf{Se}lf-supervised \textbf{R}epresentations'' is presented. LASER is based on the soft-DTW alignment loss with temporal regularisation term. Experiments are conducted with HuBERT and WavLM models and evaluated on the SUPERB benchmark for two content-related tasks: automatic speech recognition (ASR) and phoneme recognition (PR). A relative improvement of 3.7\% and 8.2\% for HuBERT, and 4.1\% and 11.7\% for WavLM are observed, for the ASR and PR tasks respectively, with only $<$ 3 hours of fine-tuning on a single GPU.

\end{abstract}

\section{Introduction}

\label{sec1}

Self-supervised learning (SSL)-based speech models are being used for full-stack speech processing \cite{wav2vec2,WavLM,distill_HuBERT}. These models are pre-trained on a large amount of unlabeled speech data with a self-supervised objective referred to as a pretext task. After pre-training, these models are then fine-tuned for downstream tasks using labeled data, and can be useful for many downstream speech applications such as automatic speech recognition (ASR), speaker identification (SID), query-by-example spoken term discovery (QbE), emotion recognition (ER), speech enhancement (SE), and speaker diarisation (SD) \cite{superb,tera}. SSL models allow for superior performance in downstream tasks compared to training models for those tasks from scratch \cite{WavLM}. This emphasises the effectiveness of SSL in utilising pre-existing knowledge to achieve better results. However, it has been observed \cite{WavLM} that the performance of SSL-based speech models on downstream tasks is often correlated with the objective of the pretext tasks. For example, models trained on top of HuBERT for speech separation achieve only marginal improvement compared with the models trained from scratch. To solve these issues, there have been a few attempts in the literature. One approach is to pre-train a model from scratch that aligns with the downstream objective. For instance, WavLM \cite{WavLM} is trained on noisy/overlapped speech to improve performance on multi-speaker tasks, such as speaker diarisation and speech separation. However, this approach is computationally expensive because the model needs to be trained from scratch.

Another approach is to use cost-effective self-supervised fine-tuning (SSFT) to fine-tune the pre-trained model based on the requirements of the downstream tasks. Then these fine-tuned models are used for further supervised fine-tuning using the labeled data for the downstream task (e.g. fine-tuning for ASR task with CTC loss on characters). The term SSFT was introduced in \cite{SPIN}, where only audio data is used for fine-tuning in self-supervised settings, rather than supervised fine-tuning with labeled data. These approaches, in general, require a marginal amount of compute cost when compared to pre-training . For instance, ContentVec \cite{ContentVec} was proposed to improve the performance on content-related tasks (ASR, PR, and QbE) by disentangling speakers, in conjunction with the pre-trained HuBERT model. However, ContentVec's efficiency is limited, requiring 19 hours of computation across 36 GPUs on top of the pre-trained HuBERT model \cite{HuBERT}. In the work \cite{SPIN}, a speaker-invariant clustering (SPIN) method for SSFT was proposed. This method clusters speech representations and performs swapped prediction between the original and speaker-perturbed utterances \cite{SPIN}. SPIN requires a compute cost of less than 1\% of ContentVec's compute cost, which demonstrates the promising aspects of SSFT. Another recent SSFT method for learning content-preserving representations is SCORE \cite{score}. SCORE employs the correspondence training method \cite{cae,simsiam}, which involves learning similar representations from two different instances of the same spoken content. Additionally, SCORE utilizes the soft-DTW loss \cite{soft-DTW} to align the representations obtained from original and perturbed speech.
Correspondence training \cite{ASRU,meghanani-hain-2024-improving} ensures that the content is preserved while  other unnecessary information such as speaker, duration, etc are marginalised. SCORE uses various speech perturbation techniques to alter the duration and  speaker information (by modifying pitch) of the utterance, which favours the learned representations to be invariant to the speaker and duration while the spoken content remains intact. Their approach involves two instances of SSL models: one trainable and the other frozen, serving to provide target outputs for the former, thus preventing representation collapse \cite{representation_collapse}. Absence of a frozen SSL model and training solely on soft-DTW alignment loss leads to a trivial solution where all embeddings cluster tightly in the embedding space \cite{LAV}, contributing to representation collapse \cite{representation_collapse,LAV}. The compute cost required by SCORE is less than $0.2$ \% of ContentVec's compute cost.

In this work, a solution to overcome representation collapse while learning content-preserving representations with the soft-DTW alignment loss is proposed. Similar to \cite{SPIN,score}, a pair of original speech and perturbed speech is generated, ensuring that the underlying content remains the same while other factors such as speaker and duration are altered. Using this pair, the soft-DTW alignment loss is applied to match the temporal sequence obtained from the SSL model along with the temporal regularisation term.
Hence, the method is named as ``LASER: \textbf{L}earning by \textbf{A}ligning \textbf{Se}lf-supervised \textbf{R}epresentations''. The temporal regularisation term assures that the embeddings are not converged to a trivial solution as described in the work \cite{LAV}, where embeddings of videos are learned with soft-DTW as alignment loss and Contrastive-Inverse Difference Moment (IDM) \cite{LAV} as temporal regularisation term. For LASER fine-tuning, the framework used in SCORE \cite{score} is adapted for speech perturbation and a modified version of Contrastive-IDM is used as temporal regularisation. After LASER fine-tuning, the models are used for supervised fine-tuning and  evaluation for two content-related tasks, ASR and PR on the Speech processing Universal PERformance Benchmark (SUPERB) \cite{superb}. LASER fine-tuned SSL models are compared with the vanilla pre-trained SSL models (HuBERT and WavLM) along with other SSFT baselines.
Later in Sec. \ref{sec4.1}, it is also demonstrated how using only soft-DTW as the alignment loss leads to representation collapse, as evidenced by reduced performance on the another content-related task QbE. This phenomenon was extensively observed and discussed in the work by \cite{LAV} in the context of learning embeddings from videos.
The main contributions of this work are as follows:
\begin{enumerate}
    \item A cost-effective content-preserving SSFT method based on the soft-DTW alignment loss with temporal regularisation is presented.
    \item Improving performance of vanilla SSL models (HuBERT and WavLM) on content-related tasks with only $<3$ hours of fine-tuning on a single GPU.
\end{enumerate}

The rest of the paper structure is as follows: Sec. \ref{sec2} introduces the proposed method; Sec. \ref{sec3} describes the experimental details; Sec. \ref{sec4} discusses the results, and finally, Sec. \ref{sec5} concludes the work.

\section{Methodology}
\label{sec2}
LASER makes use of correspondence training strategy. Self-supervised representations obtained from original speech and perturbed speech are aligned to match the common factor, i.e. content representations. In the following, the SSL model under consideration for fine-tuning (only top 2 layers of the Transformer, more details  in Sec. \ref{sec3.2}) is represented by $M_\theta$. Let $Z = \{z_1, z_2, \dots, z_m\}$ be the sequence of embedding representations for original speech, obtained from the final Transformer layer of the model $M_\theta$. Let $Z' = \{z_1',z_2',\dots, z_n'\}$ be the representations obtained for the perturbed speech. These representation are then projected to a lower-dimension and L2-normalised, denoted as $X = \{x_1, x_2, \dots, x_m\}$ and $X' = \{x_1',x_2',\dots, x_n'\}$, as shown in Fig. \ref{fig:1}.

Now these representations can be aligned with a soft-DTW loss \cite{soft-DTW,soft-dtw-cuda,diff_divergence}, a differential version of the Dynamic Time Warping (DTW) alignment metric. Soft-DTW is widely employed for time series data \cite{soft-dtw-cuda1} and is increasingly used for modelling in domains such as music \cite{mpe} and speech \cite{score}. In \cite{mpe}, it has been utilized for multi-pitch estimation, and in \cite{score}, it served as the alignment loss for content-preserving SSFT. However, optimizing for soft-DTW loss alone can result in degenerate solutions, as extensively discussed in the work \cite{LAV}. To overcome this issue, in \cite{LAV} a temporal regularisation term Contrastive-IDM was successfully employed. Contrastive-IDM optimizes for temporally disentangled representations, meaning that frames that are far apart in time are linked to spatially distant points in the embedding space, and vice versa. If $D_X \in \mathbb{R}^{m \times m}$ is the self-distance matrix of $X$ and is defined as $D_X(i, j) = ||x_i - x_j||^2$, then the  Contrastive-IDM ($f(X)$) for $X$ is defined as follows:
\begin{figure}
    \centering
    \includegraphics[width=\columnwidth]{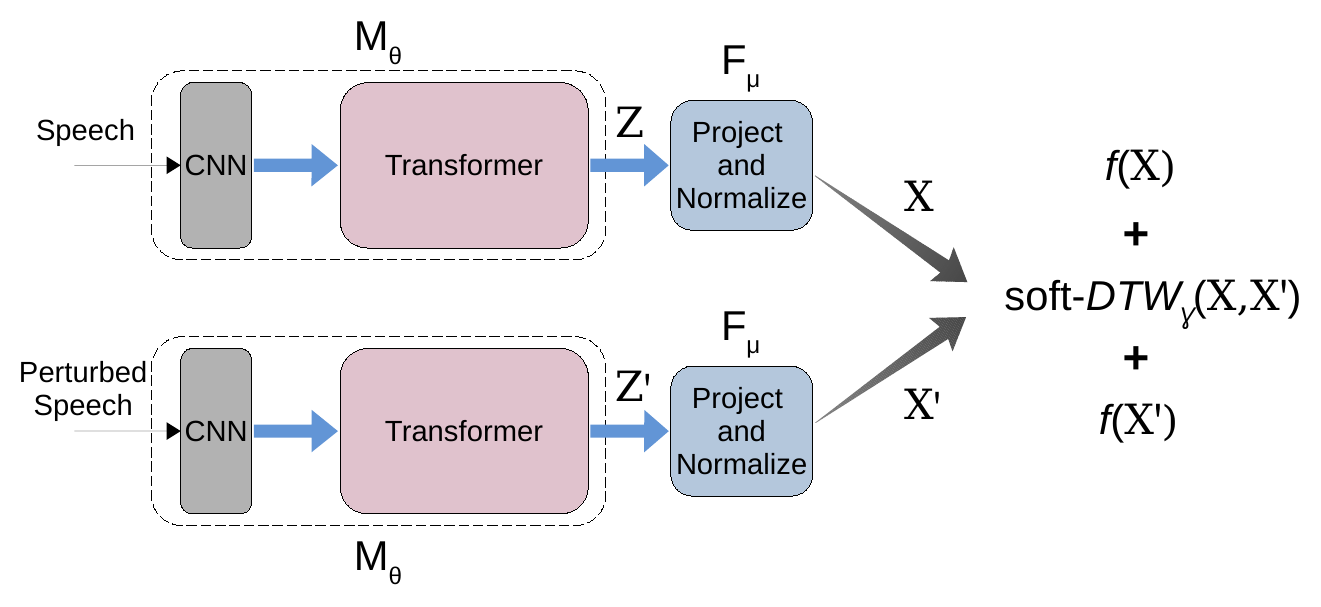}
    \caption{LASER fine-tuning approach. The loss function is computed for the representations obtained from original speech ($X$) and perturbed speech ($X'$): $L(X,X')=\text{soft-DTW}_\gamma(X,X') + \alpha(f(X) + f(X')) $. }
    \label{fig:1}
\end{figure}

\begin{equation}
\label{eq:1}
\begin{split}
    f(X) = \sum_{i=1}^{m} \sum_{j=1}^{m} y_{ij} W(i,j)  max(0,\lambda - D_X(i,j)) \\
    + (1 - y_{ij}) \frac{1}{W(i,j)} D_X(i,j),\\
    y_{ij} = \begin{cases}
        1 ,& |i - j| \geq \sigma\\
        0, & |i - j| < \sigma\\
    \end{cases}\\
\end{split}
\end{equation}

In Eq. \ref{eq:1}, $W(i,j) =(i - j)^2+1$ and $\sigma$ represents the window size for separating temporally distant frames. As seen, $f(X)$ penalizes temporally distant embeddings when the distance between them in the embedding space is smaller than the margin $\lambda$, using a scaling factor $W(i,j)$. It promotes temporally close frames to be proximate in the embedding space, with a scaling factor of $\frac{1}{W(i,j)}$. However, empirically, lower values of $\sigma$ have been found to yield optimal performance (will be discussed in more detail in Sec. \ref{sec3.4}). The value of $\sigma$ is set to 1 for the remainder of this work. When $\sigma = 1$, Eq. \ref{eq:1} can be rewritten as:

\begin{equation}
\label{eq:3}
\begin{split}
    f(X) = \sum_{i=1}^{m} \sum_{j=1}^{m} y_{ij} W(i,j)  max(0,\lambda - D_X(i,j))\\
    y_{ij} = \begin{cases}
        1 ,& i \neq j \\
        0, & i = j \\
    \end{cases}\\
\end{split}
\end{equation}

When $\sigma=1$, $f(X)$ only pushes away the embeddings of the temporally away frames in the embeddings space with weight factor $W(i,j)$, proportional to the indexes of the frame, preventing them to converge to a trivial solution. Now, the final loss, which is the sum of soft-DTW and temporal regularisation term  for both $X$ and $X'$ can be written as:

\begin{equation}
    L(X,X') = \text{soft-DTW}_\gamma(X,X') + \alpha(\frac{f(X)}{m^2}+\frac{f(X')}{n^2})    \label{eq:4}
\end{equation}

Here, $\alpha$ is the regularisation weight and $\gamma$ is the smoothing factor of soft-DTW. A normalized version of soft-DTW \cite{diff_divergence,LAV} is used in this work \footnote{\scriptsize{\url{https://github.com/trquhuytin/LAV-CVPR21}}}.  Since the embedding representations can vary in length, the temporal regularization term is normalized with $m^2$ and $n^2$ for $f(X)$ and $f(X')$ respectively. Here, $m$ and $n$ are the sequence lengths of $X$ and $X'$ respectively. The normalization term for temporal regularization involves squaring the lengths because the value of $f(X)$ exhibits quadratic growth with respect to the sequence length. The entire LASER algorithm is describe in Algo. \ref{alg:alg1}

\begin{algorithm}[t!]
\caption{LASER fine-tuning}\label{euclid}
\begin{algorithmic}[1]
\State $M_{\theta}$ = SSL model (with learnable top 2 layers only)
\State $F_{\mu}$ = Linear projection layer + L-2 Normalisation
\State Total samples in the dataset = $N_{samp}$
\State $S_i$ = i\textsuperscript{th} speech utterance
\While {Not Converged}
      \For{\texttt{i=1 to $N_{samp}$}}
        \State ${S_i}^p = SpeedPerturbation(S_i)$
        \State ${S_i}^p = PitchShift({S_i}^p)$
        \State $Z = M_{\theta}(S_i)$, $Z' = M_{\theta}({S_i}^p)$ 
    
        \State $ X = F_{\mu}(Z), X' = F_{\mu}(Z')$
        \State $ L = \text{soft-DTW}_\gamma(X,X') + \alpha(\frac{f(X)}{m^2}+\frac{f(X')}{n^2}) $
        \State Gradient computation $\frac{\partial L}{\partial \theta}$,$\frac{\partial L}{\partial \mu}$
        \State Update $\theta$ and $\mu$ to minimize $L$

      \EndFor

\EndWhile
\end{algorithmic}
\label{alg:alg1}
\end{algorithm}

\section{Experimental Setup}
\label{sec3}
\subsection{Dataset and speech perturbations}
\label{sec3.1}
Following earlier work on SSFT (SCORE \cite{score}, SPIN \cite{SPIN}), and for a fair comparison with baseline models, LibriSpeech’s \cite{libri} train-clean-100 hours of data is used for LASER fine-tuning. To obtain the required perturbed speech, the approach used in SCORE \cite{score} is adapted. SCORE employs data augmentations commonly used in ASR \cite{data_aug_asr}, such as speed perturbations and pitch shifting. Torchaudio \cite{torchaudio} is used for these perturbations, with \texttt{SpeedPerturbation} and \texttt{PitchShift} functions under \texttt{torchaudio.transforms}\footnote{\scriptsize{\url{https://pytorch.org/audio/stable/transforms.html}}}.
\subsection{Details of model ($M_\theta$)}
\label{sec3.2}
In this study, two different SSL speech models are used in experiments, namely the BASE versions of both HuBERT and WavLM, with each having roughly 95 million parameters. These models consist of multi-layer CNN models at the front-end, followed by 12 Transformer layers. The output from the final layer of Transformer block of the SSL models are 768-dimensional sequence of vectors. Consistent with the recent baselines, only the top two layers of the Transformer (11\textsuperscript{th} and 12\textsuperscript{th}) are fine-tuned ($\approx 14$ million) since most SSL models encode phonetic information in top layers \cite{layerwise_analysis}. In \cite{score,SPIN}, it has been demonstrated that training parameters in more layers does not help for content-related tasks. Additionally, training more layers defeats the purpose of this study as a cost-effective alternative. S3PRL toolkit\footnote{\scriptsize{\url{https://github.com/s3prl/s3prl}}} \cite{superb} is used for all the experiments. The code for the soft-DTW and temporal regularisation is adapted from \cite{LAV}. 

\subsection{Details of model head ($F_\mu$)}
\label{sec3.3}
The 768-dimensional representations obtained from the SSL models are projected in to lower-dimensional (256) representations with a linear projection layer and then L2-normalised \cite{SPIN,score}.

\subsection{Loss function}
\label{sec3.4}
 For soft-DTW loss, the value of $\gamma$ is taken as 0.1, a standard choice in the literature \cite{score,LAV}. To obtain the optimal values for the temporal regularisation term, a grid search over $\sigma$ (window), $\alpha$ (regularisation weight), and $\lambda$ (margin) is performed using another content-related task QbE from the SUPERB benchmark,  as it requires no extra training. For QbE, conventional supervised phoneme posteriorgram are replaced with SSL representations \cite{superb}. The evaluation on the test set is performed by running DTW on the final layer and obtain a score for each query-document pair. The best values of $\sigma$, $\alpha$, and $\lambda$
are selected based on performance of the final layer on test set from QUESST 2014 \cite{quest2014} data. The evaluation metric for QbE task on SUPERB benchmark is maximum term weighted value (MTWV in \%) \cite{superb}. The best value for $\sigma$ was found do be 1 for both HuBERT and WavLM model. For HuBERT, the best value for $\alpha$ and $\lambda$ was found to be 0.4 and 1.1. For WavLM, the best value for $\alpha$ and $\lambda$ was found to be 0.15 and 1.
\subsection{LASER fine-tuning}
\label{sec3.5}
The model is fine-tuned for 3.6K updates ($\approx$ 1 epoch) with 1k warm-up updates. One epoch was found to be sufficient for convergence and further training did not improve. An effective batch size of 8 (batch size $\times$  gradient accumulation step) is used with  AdamW \cite{adamw} optimizer and a learning rate of  $2.0e-5$. LASER fine-tuning takes $< 3$ hours on a single A100 GPU\footnote{\scriptsize{\url{https://github.com/Trikaldarshi/LASER.git}}}.

\subsection{Evaluation on the SUPERB benchmark}
\label{sec3.6}
After LASER fine-tuning, the SSL models are evaluated on the SUPERB benchmark for two content-related tasks: ASR and PR. SUPERB benchmark tasks use the weighted sum of features from all the layers, coupled with a model head. The model head itself is task dependent. The weights for the layer and model head are fine-tuned with the labels associated with the downstream task. For ASR, the model head consists of a 2-layer 1024-unit Bi-LSTM network with CTC loss on characters \cite{superb}. LibriSpeech train-clean-100/dev-clean/test-clean subsets are used for training/validation/testing for ASR \cite{superb}. The performance of ASR is evaluated without an external language model, to ensure a fair comparison between different SSL model types. For PR, the model head is a frame-wise linear transformation with CTC loss.  The same datasets as used for ASR are adopted for training/validation/testing of PR task. More details are available at SUPERB benchmark \cite{superb}. For both ASR and PR, Adam optimizer is used with learning rate of $1.0e-4$  and $5.0e-4$, respectively. All other parameter settings are available at SUPERB benchmark \cite{superb}. We run each experimental setup 5 times and report the results with mean and standard deviation.  The evaluation metric for ASR and PR are word error rate (WER in \%) and phoneme error rate (PER in \%) respectively.

\section{Results and Discussions}
\label{sec4}

\begin{table}[h!]
\caption{Processed speech during training in ``pre-training'' stage and in ``SSFT'' stage for various SSL models and their fine-tuned versions. Processed speech is defined as “training steps × effective batch duration” to quantify machine-independent training costs \cite{SPIN}.} 
\centering
\resizebox{0.9\columnwidth}{!}{%
\begin{threeparttable}
\begin{tabular}{ccc}
\specialrule{.1em}{0em}{0em} 
\multirow{2}{*}{\textbf{Model}} & \multicolumn{2}{c}{\textbf{\begin{tabular}[c]{@{}c@{}}Training\\ Processed Speech (hours)\end{tabular}}} \\ \cline{2-3}
                                & \textbf{Pre-training}                                   & \textbf{SSFT}                                                                        \\ \specialrule{.1em}{0em}{0em} 
HuBERT \cite{HuBERT}                         & 506K                                                    & 0                                                                                                                      \\
WavLM \cite{WavLM}                      & 1439K                                                   & 0                                                                                                            \\ \hline
ContentVec$_{500}$ \cite{ContentVec}              & 506K                                                    & 76K                                                                                                                 \\ \hline
HuBERT + SPIN$_{256}$ \cite{SPIN}                 & 506K                                                    & 356                                                                                                              \\
WavLM + SPIN$_{256}$ \cite{SPIN}                   & 1439K                                                   & 356                                                                                                       \\ \hline

HuBERT + SCORE \cite{score}                  & 506K                                                    & 100                                                                                                                  \\
WavLM + SCORE  \cite{score}                   & 1439K                                                   & 100                                                                                                                  \\ \hline
HuBERT + LASER                  & 506K                                                    & 100                                                                                                                  \\
WavLM + LASER                    & 1439K                                                   & 100                                                                                                                  \\ \hline

\end{tabular}%
\end{threeparttable}
}

\label{tab1}

\end{table}

\begin{table}[h!]
\caption{Results of the proposed LASER fine-tuning of HuBERT and WavLM models along with baseline methods on SUPERB benchmark. The baseline methods include the BASE version of HuBERT and WavLM models, along with SSFT based SPIN and SCORE models. The downstream tasks include ASR and PR, which are evaluated on word error rate (WER in \%) and phoneme error rate (PER in \%) respectively.} 
\centering
\resizebox{0.9\columnwidth}{!}{%
\begin{threeparttable}
\begin{tabular}{ccccc}
\specialrule{.1em}{0em}{0em} 
\multirow{2}{*}{\textbf{Model}} & \multirow{2}{*}{\textbf{\begin{tabular}[c]{@{}c@{}}ASR\\ (WER) $\downarrow$\end{tabular}}} & \multirow{2}{*}{\textbf{\begin{tabular}[c]{@{}c@{}}PR\\ (PER) $\downarrow$\end{tabular}}} \\ &
&                                                                              &                                                                                \\ \specialrule{.1em}{0em}{0em} 

HuBERT \cite{HuBERT}\tnote{$\clubsuit$}   &                     6.42 $\pm$ 0.08                                                                       & 5.02 $\pm$ 0                                                                                                                                               \\
WavLM \cite{WavLM}\tnote{$\clubsuit$}  &                        6.17 $\pm$ 0.02                                                                         & 4.85 $\pm$ 0                                                                                                                                               \\ \hline
HuBERT \cite{HuBERT}\tnote{$\diamondsuit$}                                                                 & 6.42                                                                       & 5.41                                                                                                                                            \\
WavLM \cite{WavLM}\tnote{$\diamondsuit$}    &                                                             6.21                                                                        & 4.84                                                                                                                                          \\ \hline
ContentVec$_{500}$ \cite{ContentVec}\tnote{$\diamondsuit$}    &                                                             5.70                                                                        & 4.54                                                                                                                                          \\ \hline
HuBERT + SPIN$_{256}$ \cite{SPIN}\tnote{$\diamondsuit$} &                 6.34                                                                          & 4.39                                                                                                                                              \\
WavLM + SPIN$_{256}$ \cite{SPIN}\tnote{$\diamondsuit$}   &          5.88                                                                          & 4.18                                                                         &                                                                  \\ \hline

HuBERT + SCORE \cite{score}\tnote{$\diamondsuit$} &                    6.35 $\pm$ 0.07                                                                          & 4.84 $\pm$ 0                                                                                                                                  \\
WavLM + SCORE \cite{score}\tnote{$\diamondsuit$}                    &  6.15 $\pm$ 0.04                                                                          & 4.72 $\pm$ 0                                                                                \\ \hline

HuBERT + LASER &                    6.18 $\pm$ 0.08                                                                          & 4.61 $\pm$ 0                                                                                                                                         \\
WavLM + LASER                    &  5.92 $\pm$ 0.06                                                                          &4.28 $\pm$ 0                                                                                                                                          \\ \hline

\end{tabular}%
    \smallskip
    \small
    \begin{tablenotes}
        \item[$\clubsuit$]Results when we run the SUPERB \cite{superb} recipes for HuBERT and WavLM for fair comparison.
        \item[$\diamondsuit$] Reported results are from their respective work and SUPERB leaderboard \cite{superb} as of 11/03/2024 (\url{https://superbbenchmark.org/leaderboard}).

    \end{tablenotes}
\end{threeparttable}
}

\label{tab2}

\end{table}

Table \ref{tab1} shows the amount of processed speech (training steps $\times$ effective batch duration) for various models during their pre-training and SSFT stage. Among them, LASER and SCORE are the best methods with least amount of processed speech in SSFT stage. For the performance on the downstream tasks, LASER is compared with the cost-effective baselines such as SPIN\cite{SPIN} and SCORE \cite{score} along with a stronger baseline ContentVec$_{500}$ \cite{ContentVec}, which uses 76K hours of processed speech compared to the LASER which uses only 100 hrs in SSFT stage.

Table \ref{tab2} shows the results for ASR and PR, for vanilla pre-trained SSL models (HuBERT and WavLM) along with the cost-effective baselines (SPIN and SCORE). From Table \ref{tab2}, it can be observed that LASER outperforms SCORE for both ASR and PR tasks with the same amount of processed speech in SSFT stage. When compared with the vanilla SSL models, LASER shows a relative improvement of 3.7\% and 8.2\% for HuBERT, and 4.1\% and 11.7\% for WavLM for ASR and PR tasks respectively. LASER provides competitive results with SPIN on ASR task, using only one third of the processed speech used by SPIN. For HuBERT model, LASER outperforms SPIN on ASR task. For WavLM model, there is no significant difference in WER between SPIN (5.88) and LASER (5.92 ± 0.06), assuming equivalent standard deviations, suggesting that LASER performs similar to SPIN. However, SPIN does better than LASER on the PR task for both HuBERT and WavLM models. The comparison between ContentVec and LASER reveals a trade-off between performance and cost-effectiveness. The performance gap on the PR task is marginal for ContentVec (4.54) and LASER (4.61), with LASER using only $<0.2$ \% of compute cost required by ContentVec during SSFT stage.  However, in ASR, ContentVec outperforms all baselines, including LASER, though it requires a higher computational cost of 76K hours of processed speech. This comparison underscores LASER's ability to deliver impressive results while minimizing resource expenditure, making it a compelling option for various content-related downstream applications.
\subsection{Analysing the impact of regularisation}
\label{sec4.1}
To measure the usefulness of temporal regularisation, an ablation study is conducted. HuBERT and WavLM models are LASER fine-tuned in the same manner as described in Sections \ref{sec2} and \ref{sec3}, with and without using the temporal regularization term in the loss function (Eq. \ref{eq:4}). The hyperparameter values are used as described in Sec. \ref{sec3.4}. The reported results are for the test set of the QbE task of the SUPERB benchmark by running DTW on the final layer. From Table \ref{tab3}, it can be observed that after fine-tuning only with soft-DTW, the performance of the fine-tuned HuBERT decreases from 7.19 to 5.17, and fine-tuned WavLM decreases from 9.15 to 4.44, indicating that the representations have collapsed. On the other hand, adding the regularisation term improves the performance for both models. This improvement also translates to the improvements in other content-related tasks such as ASR and PR, as shown in Table \ref{tab2}. 

\begin{table}[]
\caption{Performance on the QbE task of the SUPERB benchmark with vanilla pre-trained SSL models and their fine-tuned versions with $\text{soft-DTW}_\gamma(X,X')$ and $\text{soft-DTW}_\gamma(X,X')$ with regularisation $\alpha(f(X)/m^2+f(X')/n^2)$.}
\label{tab3}
\resizebox{\columnwidth}{!}{%
\setlength{\tabcolsep}{4pt}
\begin{tabular}{ccc}
\specialrule{.1em}{0em}{0em} 
\textbf{Loss}                                                                   & \textbf{Model}  & \textbf{QbE(MTWV) $\uparrow$} \\ 
\specialrule{.1em}{0em}{0em} 
\multirow{2}{*}{--}                                                    & HuBERT &      7.19     \\
                                                                       & WavLM  &     9.15      \\ \hline
\multirow{2}{*}{$\text{soft-DTW}_\gamma(X,X') $}                                                  & HuBERT &   5.17        \\
                                                                       & WavLM  &    4.44       \\ \hline
\multirow{2}{*}{\begin{tabular}[c]{@{}c@{}}$\text{soft-DTW}_\gamma(X,X') $\\ $+$ $\alpha(f(X)/m^2+f(X')/n^2)$\end{tabular}} & HuBERT &         8.91  \\
                                                                       & WavLM  &          9.27 \\ \hline
\end{tabular}%
}
\end{table}
\section{Conclusions}
A cost-effective SSFT method named ``LASER'' is presented for improving content representations. LASER fine-tuning is based on the correspondence training strategy with soft-DTW alignment loss and temporal regularisation. The efficacy of temporal regularization in preventing representation collapse is successfully demonstrated. LASER outperformed the recent baseline SCORE on both ASR and PR task for both HuBERT and WavLM models, which also uses soft-DTW alignment but without any regularisation term. LASER provides competitive results with SPIN with only one third of the processed speech. In future work,  we plan to use more sophisticated speech perturbation techniques. Additionally, we plan to explore LASER fine-tuning on out-of-domain data and its associated downstream tasks for acoustic model adaptation \cite{huang22b_interspeech}.
\label{sec5}

\section{Acknowledgements}

This work was supported by the Centre for Doctoral Training in Speech
and Language Technologies (SLT) and their Applications funded by UK Research and Innovation [grant number EP/S023062/1]. This work was also
funded in part by LivePerson, Inc.

\bibliographystyle{IEEEtran}
\bibliography{mybib}

\begin{thebibliography}{10}
\providecommand{\url}[1]{#1}
\csname url@samestyle\endcsname
\providecommand{\newblock}{\relax}
\providecommand{\bibinfo}[2]{#2}
\providecommand{\BIBentrySTDinterwordspacing}{\spaceskip=0pt\relax}
\providecommand{\BIBentryALTinterwordstretchfactor}{4}
\providecommand{\BIBentryALTinterwordspacing}{\spaceskip=\fontdimen2\font plus
\BIBentryALTinterwordstretchfactor\fontdimen3\font minus \fontdimen4\font\relax}
\providecommand{\BIBforeignlanguage}[2]{{%
\expandafter\ifx\csname l@#1\endcsname\relax
\typeout{** WARNING: IEEEtran.bst: No hyphenation pattern has been}%
\typeout{** loaded for the language `#1'. Using the pattern for}%
\typeout{** the default language instead.}%
\else
\language=\csname l@#1\endcsname
\fi
#2}}
\providecommand{\BIBdecl}{\relax}
\BIBdecl

\bibitem{wav2vec2}
A.~Baevski, Y.~Zhou, A.~Mohamed, and M.~Auli, ``wav2vec 2.0: A framework for self-supervised learning of speech representations,'' in \emph{Advances in Neural Information Processing Systems}, H.~Larochelle, M.~Ranzato, R.~Hadsell, M.~Balcan, and H.~Lin, Eds., vol.~33.\hskip 1em plus 0.5em minus 0.4em\relax Curran Associates, Inc., 2020, pp. 12\,449--12\,460.

\bibitem{WavLM}
S.~Chen, C.~Wang, Z.~Chen, Y.~Wu, S.~Liu, Z.~Chen, J.~Li, N.~Kanda, T.~Yoshioka, X.~Xiao, J.~Wu, L.~Zhou, S.~Ren, Y.~Qian, Y.~Qian, J.~Wu, M.~Zeng, X.~Yu, and F.~Wei, ``Wavlm: Large-scale self-supervised pre-training for full stack speech processing,'' \emph{IEEE Journal of Selected Topics in Signal Processing}, vol.~16, no.~6, pp. 1505--1518, July 2022.

\bibitem{distill_HuBERT}
H.~Chang, S.~Yang, and H.~Lee, ``Distilhubert: Speech representation learning by layer-wise distillation of hidden-unit bert,'' in \emph{Prof. of ICASSP 2022)}, 2022, pp. 7087--7091.

\bibitem{superb}
S.~Yang, P.~Chi, Y.~Chuang, C.~J. Lai, K.~Lakhotia, Y.~Y. Lin, A.~T. Liu, J.~Shi, X.~Chang, G.~Lin, T.~Huang, W.~Tseng, K.~Lee, D.~Liu, Z.~Huang, S.~Dong, S.~Li, S.~Watanabe, A.~Mohamed, and H.~Lee, ``{SUPERB: Speech Processing Universal PERformance Benchmark},'' in \emph{Proc. Interspeech 2021}, 2021, pp. 1194--1198.

\bibitem{tera}
\BIBentryALTinterwordspacing
A.~T. Liu, S.-W. Li, and H.-y. Lee, ``Tera: Self-supervised learning of transformer encoder representation for speech,'' \emph{IEEE/ACM Trans. Audio, Speech and Lang. Proc.}, vol.~29, p. 2351–2366, jul 2021. [Online]. Available: \url{https://doi.org/10.1109/TASLP.2021.3095662}
\BIBentrySTDinterwordspacing

\bibitem{SPIN}
H.~Chang, A.~H. Liu, and J.~Glass, ``{Self-supervised Fine-tuning for Improved Content Representations by Speaker-invariant Clustering},'' in \emph{Proc. INTERSPEECH 2023}, 2023, pp. 2983--2987.

\bibitem{ContentVec}
\BIBentryALTinterwordspacing
K.~Qian, Y.~Zhang, H.~Gao, J.~Ni, C.~Lai, D.~Cox, M.~Hasegawa-Johnson, and S.~Chang, ``{C}ontent{V}ec: An improved self-supervised speech representation by disentangling speakers,'' in \emph{Proceedings of the 39th International Conference on Machine Learning}, ser. Proceedings of Machine Learning Research, K.~Chaudhuri, S.~Jegelka, L.~Song, C.~Szepesvari, G.~Niu, and S.~Sabato, Eds., vol. 162.\hskip 1em plus 0.5em minus 0.4em\relax PMLR, 17--23 Jul 2022, pp. 18\,003--18\,017. [Online]. Available: \url{https://proceedings.mlr.press/v162/qian22b.html}
\BIBentrySTDinterwordspacing

\bibitem{HuBERT}
\BIBentryALTinterwordspacing
W.~Hsu, B.~Bolte, Y.~Tsai, K.~Lakhotia, R.~Salakhutdinov, and A.~Mohamed, ``Hubert: Self-supervised speech representation learning by masked prediction of hidden units,'' \emph{IEEE/ACM Trans. Audio, Speech and Lang. Proc.}, vol.~29, p. 3451–3460, oct 2021. [Online]. Available: \url{https://doi.org/10.1109/TASLP.2021.3122291}
\BIBentrySTDinterwordspacing

\bibitem{score}
A.~Meghanani and T.~Hain, ``Score: Self-supervised correspondence fine-tuning for improved content representations,'' in \emph{ICASSP 2024 - 2024 IEEE International Conference on Acoustics, Speech and Signal Processing (ICASSP)}, 2024.

\bibitem{cae}
H.~Kamper, ``Truly unsupervised acoustic word embeddings using weak top-down constraints in encoder-decoder models,'' \emph{ICASSP 2019 - 2019 IEEE International Conference on Acoustics, Speech and Signal Processing (ICASSP)}, pp. 6535--3539, 2019.

\bibitem{simsiam}
X.~Chen and K.~He, ``Exploring simple siamese representation learning,'' in \emph{2021 IEEE/CVF Conference on Computer Vision and Pattern Recognition (CVPR)}, 2021, pp. 15\,745--15\,753.

\bibitem{soft-DTW}
\BIBentryALTinterwordspacing
M.~Cuturi and M.~Blondel, ``Soft-{DTW}: a differentiable loss function for time-series,'' in \emph{Proceedings of the 34th International Conference on Machine Learning}, ser. Proceedings of Machine Learning Research, D.~Precup and Y.~W. Teh, Eds., vol.~70.\hskip 1em plus 0.5em minus 0.4em\relax PMLR, 06--11 Aug 2017, pp. 894--903. [Online]. Available: \url{https://proceedings.mlr.press/v70/cuturi17a.html}
\BIBentrySTDinterwordspacing

\bibitem{ASRU}
A.~Meghanani and T.~H., ``Deriving translational acoustic sub-word embeddings,'' in \emph{2023 IEEE Automatic Speech Recognition and Understanding Workshop (ASRU)}, 2023, pp. 1--8.

\bibitem{meghanani-hain-2024-improving}
\BIBentryALTinterwordspacing
A.~Meghanani and T.~Hain, ``Improving acoustic word embeddings through correspondence training of self-supervised speech representations,'' in \emph{Proceedings of the 18th Conference of the European Chapter of the Association for Computational Linguistics (Volume 1: Long Papers)}, Y.~Graham and M.~Purver, Eds.\hskip 1em plus 0.5em minus 0.4em\relax Association for Computational Linguistics, Mar. 2024, pp. 1959--1967. [Online]. Available: \url{https://aclanthology.org/2024.eacl-long.118}
\BIBentrySTDinterwordspacing

\bibitem{representation_collapse}
\BIBentryALTinterwordspacing
A.~Aghajanyan, A.~Shrivastava, A.~Gupta, N.~Goyal, L.~Zettlemoyer, and S.~Gupta, ``Better fine-tuning by reducing representational collapse,'' \emph{CoRR}, vol. abs/2008.03156, 2020. [Online]. Available: \url{https://arxiv.org/abs/2008.03156}
\BIBentrySTDinterwordspacing

\bibitem{LAV}
S.~Haresh, S.~Kumar, H.~Coskun, S.~N. Syed, A.~Konin, M.~Z. Zia, and Q.-H. Tran, ``Learning by aligning videos in time,'' in \emph{2021 IEEE/CVF Conference on Computer Vision and Pattern Recognition (CVPR)}, 2021, pp. 5544--5554.

\bibitem{soft-dtw-cuda}
M.~Maghoumi, E.~M. Taranta, and J.~LaViola, ``Deepnag: Deep non-adversarial gesture generation,'' in \emph{26th International Conference on Intelligent User Interfaces}, 2021, pp. 213--223.

\bibitem{diff_divergence}
\BIBentryALTinterwordspacing
M.~Blondel, A.~Mensch, and J.~Vert, ``Differentiable divergences between time series,'' in \emph{Proc. of AIStat}, ser. Proceedings of Machine Learning Research, vol. 130.\hskip 1em plus 0.5em minus 0.4em\relax PMLR, 13--15 Apr 2021, pp. 3853--3861. [Online]. Available: \url{https://proceedings.mlr.press/v130/blondel21a.html}
\BIBentrySTDinterwordspacing

\bibitem{soft-dtw-cuda1}
M.~Maghoumi, ``{Deep Recurrent Networks for Gesture Recognition and Synthesis},'' Ph.D. dissertation, University of Central Florida Orlando, Florida, 2020.

\bibitem{mpe}
M.~Krause, C.~Weiß, and M.~Müller, ``Soft dynamic time warping for multi-pitch estimation and beyond,'' in \emph{ICASSP 2023 - 2023 IEEE International Conference on Acoustics, Speech and Signal Processing (ICASSP)}, 2023, pp. 1--5.

\bibitem{libri}
V.~Panayotov, G.~Chen, D.~Povey, and S.~Khudanpur, ``Librispeech: An asr corpus based on public domain audio books,'' in \emph{2015 IEEE International Conference on Acoustics, Speech and Signal Processing (ICASSP)}, 2015, pp. 5206--5210.

\bibitem{data_aug_asr}
T.~Ko, V.~Peddinti, P.~D, and S.~Khudanpur, ``{Audio augmentation for speech recognition},'' in \emph{Proc. Interspeech 2015}, 2015, pp. 3586--3589.

\bibitem{torchaudio}
Y.~Yang, M.~Hira, Z.~Ni, A.~Astafurov, C.~Chen, C.~Puhrsch, D.~Pollack, D.~Genzel, D.~Greenberg, E.~Z. Yang, J.~Lian, J.~Hwang, J.~Chen, P.~Goldsborough, S.~Narenthiran, S.~Watanabe, S.~Chintala, and V.~Quenneville-Bélair, ``Torchaudio: Building blocks for audio and speech processing,'' in \emph{ICASSP 2022 - 2022 IEEE International Conference on Acoustics, Speech and Signal Processing (ICASSP)}, 2022, pp. 6982--6986.

\bibitem{layerwise_analysis}
A.~Pasad, J.~Chou, and K.~Livescu, ``Layer-wise analysis of a self-supervised speech representation model,'' in \emph{2021 IEEE Automatic Speech Recognition and Understanding Workshop (ASRU)}, 2021, pp. 914--921.

\bibitem{quest2014}
X.~Anguera, L.~Rodriguez-Fuentes, A.~Buzo, F.~Metze, I.~Szöke, and M.~Penagarikano, ``Quesst2014: Evaluating query-by-example speech search in a zero-resource setting with real-life queries,'' in \emph{2015 IEEE International Conference on Acoustics, Speech and Signal Processing (ICASSP)}, 2015, pp. 5833--5837.

\bibitem{adamw}
\BIBentryALTinterwordspacing
I.~Loshchilov and F.~Hutter, ``Decoupled weight decay regularization,'' in \emph{International Conference on Learning Representations}, 2019. [Online]. Available: \url{https://openreview.net/forum?id=Bkg6RiCqY7}
\BIBentrySTDinterwordspacing

\bibitem{huang22b_interspeech}
K.~P. Huang, Y.~Fu, Y.~Zhang, and H.~Lee, ``{Improving Distortion Robustness of Self-supervised Speech Processing Tasks with Domain Adaptation},'' in \emph{Proc. Interspeech 2022}, 2022, pp. 2193--2197.

\end{thebibliography}

\end{document}